\documentclass[review]{elsarticle}

\usepackage{lineno,hyperref}
\usepackage{multirow}
\usepackage{multicol}
\usepackage{bm}
\modulolinenumbers[5]

\journal{Journal of \LaTeX\ Templates}

\begin{document}

\begin{frontmatter}

\title{EFSIS: Ensemble Feature Selection Integrating Stability}


\author[mymainaddress]{Xiaokang Zhang}
\ead{Xiaokang.Zhang@uib.no}

\author[mymainaddress]{Inge Jonassen\corref{mycorrespondingauthor}}
\cortext[mycorrespondingauthor]{Corresponding author}
\ead{Inge.Jonassen@uib.no}

\address[mymainaddress]{Computational Biology Unit, Department of Informatics, University of Bergen, Norway}

\begin{abstract}
Ensemble learning that can be used to combine the predictions from multiple learners has been widely applied in pattern recognition, and has been reported to be more robust and accurate than the individual learners. This ensemble logic has recently also been more applied in feature selection. There are basically two strategies for ensemble feature selection, namely data perturbation and function perturbation. Data perturbation performs feature selection on data subsets sampled from the original dataset and then selects the features consistently ranked highly across those data subsets. This has been found to improve both the stability of the selector and the prediction accuracy for a classifier. Function perturbation frees the user from having to decide on the most appropriate selector for any given situation and works by aggregating multiple selectors. This has been found to maintain or improve classification performance. Here we propose a framework, EFSIS, combining these two strategies. Empirical results indicate that EFSIS gives both high prediction accuracy and stability. 
\end{abstract}


\end{frontmatter}


\section{Introduction}

Feature selection is a crucial technique in machine learning. It is widely used in many fields to help to find the most important features from the whole feature set. In classification tasks, feature selection can help to improve the prediction accuracy by removing the noisy features and avoiding overfitting. But feature selection can also be very challenging, especially when there is a large number of features (high-dimension) and few training samples. In such cases a small change in the samples used as training set, can sometimes lead to a large change in the set of selected features. The ability of a feature selection method to give a consistent set of features  when the training data changes, is called stability. So a good feature selection method should enable the chosen classifier to obtain high prediction accuracy and also be stable to provide similar selected feature subsets.

Many techniques for feature selection have been proposed, and among them, ensemble feature selection has drawn more attention recently. It has been observed that the ensemble of multiple prediction models can achieve higher stability and prediction accuracy \cite{Sagi2018}. The ensemble logic was also applied to feature selection in recent years.

Ensemble feature selection methods can mainly be divided into two categories: data perturbation and function perturbation \cite{He2010}. 

In data perturbation (sometimes referred to as homogeneous ensemble approach), feature selection is performed on several subsets of the samples, each analysis generating potentially different feature subsets. In this case the same feature selection method is used to analyze all subsets. The resulting feature subsets are then aggregated into one final feature subset \cite{Davis2006, Bach2008, Abeel2010, Seijo-Pardo2017, Pes2017a}. \cite{Pes2017a} showed that data perturbation can improve the stability of the original feature selection method.

Function perturbation (also known as heterogeneous ensemble approach) combines the outputs from several feature selection methods applied on the same training set to take account of the strengths and weaknesses of each method \cite{Tan2009, BenBrahim2013a, Ahmed2014, Seijo-Pardo2017}. According to the literature, function perturbation can maintain or improve classification performance. More importantly, it can free the researchers from having to choose one feature selection method to be used in a specific setting. However, we have not seen any study of the stability of function-perturbation based methods for feature selection.

Since data perturbation has been shown to improve stability and function perturbation to improve prediction accuracy, we wanted to investigate if it would be possible to use both in combination and achieve both higher stability and improved prediction accuracy. For this purpose, we propose the EFSIS (Ensemble Feature Selection Integrating Stability) framework combining both approaches and using the stability of each feature selection method to perform a “weighted voting” in order to obtain the consensus feature set. The source code is available on GitHub (https://github.com/zhxiaokang/EFSIS).

As benchmarks for our experiments, we tested our method on six cancer datasets coming from microarray experiments. To better understand its performance, we compared EFSIS with each of the methods aggregated in EFSIS and also with basic function perturbation. We recorded each method's performance in terms of prediction accuracy and stability on each of the six datasets.

The rest of the article is organized as follows: Section 2 describes the proposed EFSIS framework, along with basic function perturbation, the aggregated individual methods, and the evaluating metrics on stability and prediction accuracy; Section 3 introduces the experimental study, including experimental settings and results; Section 4 discusses the experiments and concludes the work.

\section{Methods}
\subsection{Methodology of EFSIS}
Our proposed ensemble feature selection approach includes two phases: data perturbation and function perturbation. The framework is illustrated in Figure 1.

\begin{figure}[ht]
\centering\includegraphics[width=1\linewidth]{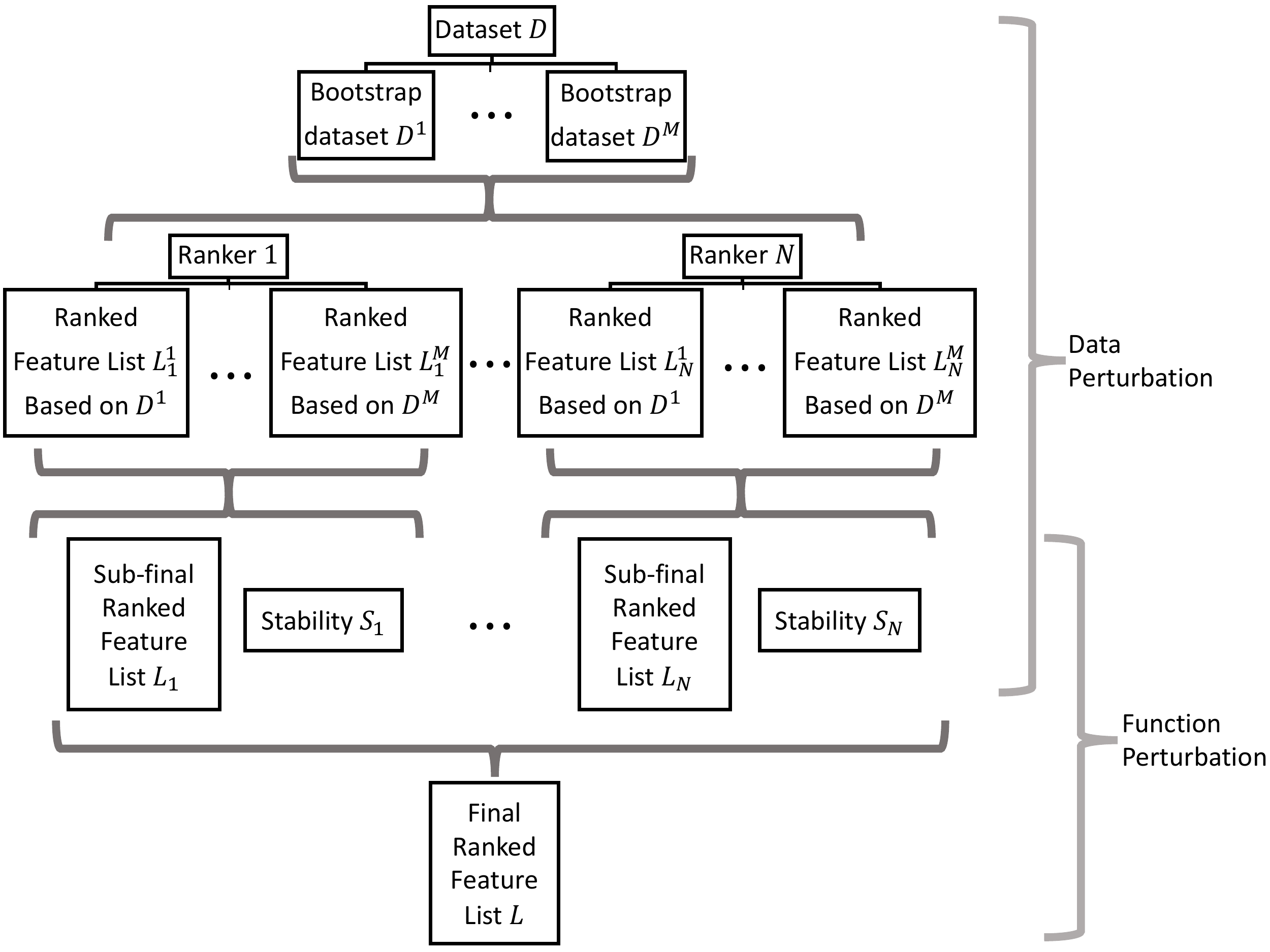}
\caption{The framework of EFSIS}
\end{figure}

Given the original dataset $D$, we use bootstrapping to get $M$ perturbed variants of $D$ ($\{D^1, ...D^m, ...D^M\}$) for the dataset $D$ with $p$ samples: we randomly draw $p$ samples from $D$ with replacement, allowing some samples to be picked multiple times while some samples may be absent in $D^m$. Each bootstrap dataset $D^m$ is then passed to each of the included individual feature selection methods, each performing a ranking of all the features based on how well they distinguish samples from different groups. For simplicity, in the following, we call each feature selection method a \emph{ranker}. 

In the first phase which is data perturbation, let us take one ranker, ranker $n$ ($n\in\{1, ... N\}$), as a general representative to explain the idea of data perturbation. Ranker $n$ will rank the features based on the bootstrap datasets. Corresponding to each bootstrap dataset, one ranked list will be generated. Therefore, each ranker will end up with $M$ ranked lists $\{L_n^1, ...L_n^m, ...L_n^M\}$. With an aggregation strategy (Equation (2) in Subsection 2.3), the $M$ lists can then be combined into one list ($L_n$). In addition to $L_n$, a “side product”, the stability of ranker $n$, that we will denote as $S_n$, can be calculated using the stability definition described in Subsection 2.2: with a pre-defined threshold $t$, the top $t$ features in $L_n^m$ will be picked to constitute a feature subset, and then the $M$ feature subsets will be used to calculate the stability of ranker $n$. The data perturbation procedure above will be applied to all $N$ rankers to generate $N$ sub-final ranked feature lists $\{L_1, ... L_N\}$.

In the phase of function perturbation, another aggregation strategy which integrates the stability of the rankers (Equation (3) in Subsection 2.3) combines those $N$ sub-final ranked feature lists into one final list $L$. The top $t$ features are kept as the selected important features by EFSIS.

\subsection{Stability}
A stable feature selection method should give similar feature subsets even given varying samples. We use the similarity between feature subsets derived from different sample sets to measure the stability of the corresponding feature selection method. We used the stability definition proposed by \cite{Davis2006}:
\begin{equation}
    S_n=\frac{\sum_{f\in{F}}(freq(f)/M)}{|F|}
\end{equation}
Where $S_n$ is the stability of a given feature selection method $n$; $M$ is the number of feature subsets analyzed; $F$ is the set of features that appear in at least one of the $M$ subsets and $|F|$ indicates the cardinality of $F$; $freq(f)$ is the frequency of feature $f\in{F}$ that appears in those $M$ subsets.

\subsection{Aggregation strategies}
There are two aggregations in the EFSIS paradigm shown in Figure 1. To aggregate the rankings of one ranker from different bootstrap datasets, we use the product of ranking positions of one feature in different ranked lists as its aggregated ranking score \cite{Breitling2004}. The ranking score of a feature $f$ from ranker $n$ can be calculated as follows:
\begin{equation}
    R_{f,n}=\prod_{m=1}^{M}R_{f,n}^m
\end{equation}
where $R_{f,n}^m$ is the rank of feature $f$ from ranker $n$ on bootstrap set $m$. 
Based on this score, an aggregated ranked feature list $L_n$ for ranker $n$ can be obtained. 

The function perturbation phase also applies the product aggregation strategy, but the stability of every ranker is used as its weight. The ranking score of a feature $f$ in the final ranked list can be calculated as follows:
\begin{equation}
    R_{f}=\prod_{n=1}^{N}(R_{f,n})^{(1-S_{n})}
\end{equation}
where $1-S_n$ is defined as the weight of ranker $n$, so that a more stable ranker is assigned a higher weight. Ranking the features based on this score, we get the final ranked list.

In basic function perturbation, each ranker will rank the features based on the original dataset $D$. Afterwards, it will apply the same product aggregation strategy, aggregating the rankings from different rankers in a similar way as EFSIS does in the second phase, except that there is no weight for each ranker ($S_n=0$ in Equation (3)).

\subsection{Individual feature selection methods}
In general, there are three categories of feature selection methods: filter methods which rank the features only based on their correlation with the targeted classes, wrapper methods which use an objective function (can be the prediction accuracy obtained by the classifier using the selected features) to evaluate features, and embedded methods where the classifier itself performs feature selection. Since one motivation of the ensemble approach is to make the method as generalizable as possible, we want to make the proposed method classifier-independent. Therefore we consider only filter methods in this context.

In our experiment, we considered four individual feature selection methods which are based on different sets of assumptions, to make our ensemble approach diverse and generalized. In particular, we employed both univariate techniques which treat the features as independent from each other and multivariate techniques which take the interaction between features into consideration.

As representatives of univariate techniques, we used:

\begin{itemize}
\item Significance Analysis of Microarrays (SAM) that was originally designed to identify genes with significantly differential expression in microarray experiments \cite{GossTusher2001}. It assigns a score to each gene based on the change in gene expression relative to the standard deviation of repeated experiments. 
\item Information gain which is one of the most popular univariate methods \cite{Witten2016}. It evaluates each feature based on the entropy concept from information theory.
\end{itemize}

As representatives of multivariate techniques, we applied:

\begin{itemize}
\item The Characteristic Direction method (GeoDE) which is a geometrical multivariate approach \cite{Clark2014a}. It defines a separating hyperplane using linear discriminant analysis to characterize the differential expression of microarray or RNA-Seq data.
\item ReliefF \cite{Kononenko1994} is an extension of the original Relief algorithm \cite{Kira1992a, Kira1992b} that evaluates a feature according to how well it can distinguish among instances that are near to each other. Compared to Relief, ReliefF is more robust to noisy and incomplete datasets. 
\end{itemize}

\subsection{Classification algorithm}
In evaluating the predictive performance of the selected feature subsets, we applied the classification algorithm Support Vector Machine (SVM) \cite{Cortes1995} to learn a classifier based on the selected feature subsets. Provided with a training dataset of samples marked with group labels (samples are characterized by the selected features), SVM will learn an optimal hyperplane separating the samples from different groups. And the optimal hyperplane will be used to predict the labels of the samples from test set. A prediction accuracy can be calculated comparing the predicted labels with the true labels. A better feature subset will enable the SVM to achieve a higher prediction accuracy. For simplicity, we chose a linear kernel for SVM and we used Area Under Curve (AUC) \cite{Fawcett2006} to summarize the obtained prediction accuracy.

\section{Experimental study}
\subsection{Datasets}
The proposed ensemble method was tested on six gene expression datasets produced using microarrays to study different forms of cancer (datasets were collected by \cite{Lazzarini2017}). The main characteristics of the datasets, including numbers of features and samples, are given in Table 1. Feature selection can provide valuable information in such applications. The selected features can be regarded as biomarkers and they reflect characteristics of the studied cancer forms and can help to classify the patients. Feature selection can allow the cancer researcher or clinician to focus on a small number of biomarkers instead of thousands of features, which can save lots of money and time for further studies. Biomarkers can also help to improve the understanding of the cancer forms on a molecular level. 

\begin{table}[ht]
\centering
\begin{tabular}{l r r l}
\hline
\textbf{Name} & \textbf{Features} & \textbf{Samples} & \textbf{Refs}\\
\hline
AML & 12 625 & 54 & \cite{Yagi2003} \\
CNS & 7 129 & 60 & \cite{Pomeroy2002} \\
DLBCL & 7 129 & 77 & \cite{Shipp2002} \\
Prostate & 12 600 & 102 & \cite{Singh2002} \\
Leukemia & 7 129 & 72 & \cite{Golub1999} \\
ColonBreast & 22 283 & 52 & \cite{Chowdary2006} \\
\hline
\end{tabular}
\caption{Datasets used in the experiments}
\end{table}

\subsection{Experimental procedure and settings}
To evaluate the performance of EFSIS, it was compared with the aggregated individual rankers and the corresponding basic function perturbation aggregating the same four rankers. The performance was evaluated in two aspects: prediction accuracy and stability. Both prediction accuracy and stability depend on how many features are to be selected and used for classification (denoted $t$), hence we performed the assessment with a range of values for $t$. 

In order to obtain an unbiased estimation of performance, we performed the experiments using a ten-fold cross-validation scheme \cite{Jensen2000, Tsamardinos2018}. Thus, we obtained 10 selected feature subsets for each  pre-defined threshold $t$, for each dataset and for each ranker. By doing classification analysis with those 10 feature subsets, we obtained 10 prediction accuracy scores. At the same time, by calculating the similarity of those 10 feature subsets using Equation (1), we obtained an estimate of the stability of the corresponding ranker.

Considering the highly variable number of  features in each dataset (as shown in Table 1), instead of using an absolute number of features $t$, we used a percentage of the original number of features. We explored a range of values from 0.3\% to 5\%.

The main parameters for EFSIS are the number of bootstrap datasets $M$ and number of rankers $N$. $M$ was chosen based on the recommendation in \cite{Pes2017a} ($M=50$). In our analysis, $N=4$, the rankers are described in Subsection 2.4. The competitors of EFSIS would therefore be the four individual rankers and the basic function perturbation of the same four rankers.

\subsection{Experimental results of predictive performance}

\begin{table}[ht]
\centering
\resizebox{\columnwidth}{!}{\begin{tabular}{*{10}{c|}c}
\hline
\multirow{2}{*}{\textbf{Dataset}} & \multirow{2}{*}{\textbf{Ranker}} & \multicolumn{9}{c}{\textbf{Percentage of selected features (\%)}}\\
\cline{3-11}
 & & 0.3 & 0.5 & 0.7 & 1 & 1.5 & 2 & 3 & 4 & 5 \\
 \hline
\multirow{6}*{AML} & SAM & $\bm{0.69\pm0.17^{\ast}}$ & $0.73\pm0.16$ & $0.73\pm0.20$ & $0.76\pm0.14$ & $0.78\pm0.17$ & $0.75\pm0.18$ & $\bm{0.74\pm0.20^{\ast}}$ & $0.76\pm0.17$ & $0.77\pm0.16$ \\
\cline{2-11}
 & GeoDE & $0.74\pm0.16$ & $0.69\pm0.25$ & $0.76\pm0.20^{\dagger}$ & $\bm{0.76\pm0.16^{\ast}}$ & $0.80\pm0.16^{\dagger}$ & $0.80\pm0.18^{\dagger}$ & $0.84\pm0.15^{\dagger}$ & $0.79\pm0.18$ & $0.79\pm0.22$ \\
 \cline{2-11}
 & ReliefF & $0.76\pm0.21$ & $0.73\pm0.15$ & $0.68\pm0.21$ & $\bm{0.69\pm0.14^{\ast}}$ & $0.75\pm0.15$ & $0.76\pm0.18$ & $\bm{0.72\pm0.14^{\ast}}$ & $0.74\pm0.18$ & $0.76\pm0.15$ \\
 \cline{2-11}
 & Info\_Gain & $0.81\pm0.16^{\dagger}$ & $0.75\pm0.18^{\dagger}$ & $0.74\pm0.16$ & $\bm{0.73\pm0.17^{\ast}}$ & $0.79\pm0.14$ & $0.76\pm0.17$ & $0.77\pm0.17$ & $0.79\pm0.16$ & $0.80\pm0.17$ \\
 \cline{2-11}
 & Func\_Pert & $0.75\pm0.16$ & $0.73\pm0.23$ & $0.74\pm0.15$ & $0.81\pm0.14^{\dagger}$ & $0.79\pm0.13$ & $0.79\pm0.17$ & $\bm{0.75\pm0.21^{\ast}}$ & $0.80\pm0.18^{\dagger}$ & $0.81\pm0.14^{\dagger}$ \\
 \cline{2-11}
 & EFSIS & $0.73\pm0.20$ & $0.74\pm0.17$ & $0.72\pm0.22$ & $0.75\pm0.17$ & $0.72\pm0.18$ & $0.73\pm0.19$ & $\bm{0.77\pm0.16^{\ast}}$ & $0.75\pm0.19$ & $0.76\pm0.17$ \\
\hline
\multirow{6}*{CNS} & SAM & $0.72\pm0.22$ & $\bm{0.69\pm0.22^{\ast}}$ & $0.71\pm0.20$ & $0.72\pm0.17$ & $\bm{0.71\pm0.21^{\ast}}$ & $\bm{0.72\pm0.17^{\ast}}$ & $\bm{0.73\pm0.18^{\ast}}$ & $\bm{0.69\pm0.19^{\ast}}$ & $\bm{0.73\pm0.18^{\ast}}$ \\
\cline{2-11}
 & GeoDE & $0.63\pm0.16$ & $0.76\pm0.08$ & $0.81\pm0.16^{\dagger}$ & $0.82\pm0.13^{\dagger}$ & $0.82\pm0.18^{\dagger}$ & $0.88\pm0.17^{\dagger}$ & $0.89\pm0.16^{\dagger}$ & $0.88\pm0.14^{\dagger}$ & $0.90\pm0.14^{\dagger}$ \\
 \cline{2-11}
 & ReliefF & $0.69\pm0.18$ & $0.72\pm0.14$ & $0.75\pm0.16$ & $\bm{0.68\pm0.15^{\ast}}$ & $0.74\pm0.19$ & $\bm{0.70\pm0.19^{\ast}}$ & $\bm{0.75\pm0.16^{\ast}}$ & $0.79\pm0.21$ & $\bm{0.73\pm0.15^{\ast}}$ \\
 \cline{2-11}
 & Info\_Gain & $0.69\pm0.17$ & $0.78\pm0.18^{\dagger}$ & $0.76\pm0.19$ & $0.71\pm0.19$ & $\bm{0.65\pm0.17^{\ast}}$ & $\bm{0.70\pm0.13^{\ast}}$ & $\bm{0.66\pm0.18^{\ast}}$ & $\bm{0.71\pm0.16^{\ast}}$ & $\bm{0.78\pm0.15^{\ast}}$ \\
 \cline{2-11}
 & Func\_Pert & $0.72\pm0.12$ & $0.77\pm0.18$ & $0.68\pm0.21$ & $\bm{0.68\pm0.21^{\ast}}$ & $0.77\pm0.15$ & $0.80\pm0.21$ & $\bm{0.80\pm0.11^{\ast}}$ & $0.80\pm0.13$ & $\bm{0.80\pm0.16^{\ast}}$ \\
 \cline{2-11}
 & EFSIS & $0.74\pm0.22^{\dagger}$ & $0.69\pm0.16$ & $0.68\pm0.19$ & $0.75\pm0.15$ & $0.79\pm0.16$ & $0.83\pm0.14$ & $0.82\pm0.14$ & $\bm{0.78\pm0.11^{\ast}}$ & $\bm{0.79\pm0.15^{\ast}}$ \\
\hline
\multirow{6}*{DLBCL} & SAM & $0.91\pm0.13$ & $0.90\pm0.12$ & $0.96\pm0.08$ & $0.96\pm0.06$ & $0.97\pm0.07$ & $0.97\pm0.07$ & $0.95\pm0.11$ & $0.94\pm0.11$ & $0.97\pm0.07^{\dagger}$ \\
\cline{2-11}
 & GeoDE & $\bm{0.86\pm0.10^{\ast}}$ & $\bm{0.87\pm0.10^{\ast}}$ & $\bm{0.86\pm0.12^{\ast}}$ & $\bm{0.89\pm0.10^{\ast}}$ & $\bm{0.88\pm0.16^{\ast}}$ & $\bm{0.86\pm0.22^{\ast}}$ & $\bm{0.85\pm0.22^{\ast}}$ & $\bm{0.89\pm0.11^{\ast}}$ & $0.92\pm0.10$ \\
 \cline{2-11}
 & ReliefF & $0.96\pm0.08^{\dagger}$ & $0.94\pm0.11$ & $0.99\pm0.03^{\dagger}$ & $0.96\pm0.09$ & $0.98\pm0.06$ & $0.94\pm0.10$ & $0.99\pm0.03^{\dagger}$ & $0.99\pm0.03^{\dagger}$ & $0.97\pm0.07^{\dagger}$ \\
 \cline{2-11}
 & Info\_Gain & $0.95\pm0.11$ & $0.95\pm0.09^{\dagger}$ & $0.95\pm0.11$ & $0.96\pm0.08$ & $0.96\pm0.08$ & $0.96\pm0.08$ & $0.96\pm0.08$ & $0.97\pm0.08$ & $0.96\pm0.06$ \\
 \cline{2-11}
 & Func\_Pert & $0.91\pm0.12$ & $0.92\pm0.09$ & $0.96\pm0.08$ & $0.96\pm0.06$ & $0.98\pm0.05^{\dagger}$ & $0.98\pm0.05^{\dagger}$ & $0.96\pm0.07$ & $0.94\pm0.10$ & $0.93\pm0.11$ \\
 \cline{2-11}
 & EFSIS & $0.92\pm0.10$ & $0.94\pm0.08$ & $\bm{0.93\pm0.10^{\ast}}$ & $0.97\pm0.06^{\dagger}$ & $0.97\pm0.06$ & $0.97\pm0.06$ & $0.96\pm0.07$ & $0.97\pm0.07$ & $0.97\pm0.07^{\dagger}$ \\
\hline
\multirow{6}*{Prostate} & SAM & $0.95\pm0.08^{\dagger}$ & $0.95\pm0.08$ & $0.95\pm0.08$ & $0.96\pm0.06^{\dagger}$ & $0.96\pm0.07$ & $0.95\pm0.07$ & $0.96\pm0.07$ & $0.96\pm0.07$ & $0.96\pm0.07^{\dagger}$ \\
\cline{2-11}
 & GeoDE & $0.90\pm0.15$ & $0.93\pm0.09$ & $0.94\pm0.09$ & $0.95\pm0.08$ & $0.95\pm0.09$ & $\bm{0.94\pm0.08^{\ast}}$ & $0.95\pm0.06$ & $0.95\pm0.06$ & $0.96\pm0.06$ \\
 \cline{2-11}
 & ReliefF & $0.94\pm0.08$ & $0.96\pm0.08^{\dagger}$ & $0.96\pm0.06^{\dagger}$ & $0.94\pm0.10$ & $0.97\pm0.04$ & $0.96\pm0.06$ & $0.94\pm0.08$ & $0.96\pm0.07$ & $0.94\pm0.09$ \\
 \cline{2-11}
 & Info\_Gain & $0.94\pm0.11$ & $0.94\pm0.10$ & $0.94\pm0.10$ & $0.95\pm0.09$ & $\bm{0.95\pm0.09^{\ast}}$ & $0.96\pm0.07$ & $0.97\pm0.06^{\dagger}$ & $0.97\pm0.06^{\dagger}$ & $0.96\pm0.08$ \\
 \cline{2-11}
 & Func\_Pert & $0.95\pm0.09$ & $0.94\pm0.10$ & $0.95\pm0.10$ & $0.95\pm0.09$ & $0.96\pm0.06$ & $0.96\pm0.07$ & $0.96\pm0.06$ & $0.95\pm0.08$ & $0.95\pm0.09$ \\
 \cline{2-11}
 & EFSIS & $0.95\pm0.09$ & $0.94\pm0.10$ & $0.95\pm0.09$ & $0.95\pm0.08$ & $0.97\pm0.07^{\dagger}$ & $0.97\pm0.07^{\dagger}$ & $0.95\pm0.08$ & $0.94\pm0.09$ & $0.94\pm0.09$ \\
\hline
\multirow{6}*{Leukemia} & SAM & $0.99\pm0.04$ & $0.98\pm0.05$ & $0.99\pm0.02^{\dagger}$ & $0.99\pm0.02^{\dagger}$ & $0.99\pm0.02^{\dagger}$ & $0.99\pm0.02^{\dagger}$ & $0.99\pm0.02^{\dagger}$ & $0.99\pm0.02^{\dagger}$ & $0.99\pm0.02^{\dagger}$ \\
\cline{2-11}
 & GeoDE & $0.98\pm0.05$ & $0.99\pm0.02^{\dagger}$ & $0.99\pm0.04$ & $0.99\pm0.02^{\dagger}$ & $0.99\pm0.02^{\dagger}$ & $0.99\pm0.02^{\dagger}$ & $0.99\pm0.02^{\dagger}$ & $0.99\pm0.02^{\dagger}$ & $0.99\pm0.02^{\dagger}$ \\
 \cline{2-11}
 & ReliefF & $0.99\pm0.02^{\dagger}$ & $0.99\pm0.02^{\dagger}$ & $0.99\pm0.04$ & $0.99\pm0.04$ & $0.99\pm0.02^{\dagger}$ & $0.99\pm0.04$ & $0.98\pm0.04$ & $0.98\pm0.05$ & $0.98\pm0.04$ \\
 \cline{2-11}
 & Info\_Gain & $0.99\pm0.02^{\dagger}$ & $0.99\pm0.02^{\dagger}$ & $0.99\pm0.02^{\dagger}$ & $0.99\pm0.02^{\dagger}$ & $0.99\pm0.02^{\dagger}$ & $0.99\pm0.02^{\dagger}$ & $0.99\pm0.02^{\dagger}$ & $0.99\pm0.02^{\dagger}$ & $0.99\pm0.02^{\dagger}$ \\
 \cline{2-11}
 & Func\_Pert & $0.97\pm0.08$ & $0.99\pm0.02^{\dagger}$ & $0.99\pm0.02^{\dagger}$ & $0.99\pm0.02^{\dagger}$ & $0.99\pm0.02^{\dagger}$ & $0.99\pm0.02^{\dagger}$ & $0.99\pm0.02^{\dagger}$ & $0.99\pm0.02^{\dagger}$ & $0.99\pm0.02^{\dagger}$ \\
 \cline{2-11}
 & EFSIS & $0.99\pm0.02^{\dagger}$ & $0.99\pm0.02^{\dagger}$ & $0.99\pm0.02^{\dagger}$ & $0.99\pm0.02^{\dagger}$ & $0.99\pm0.02^{\dagger}$ & $0.99\pm0.02^{\dagger}$ & $0.99\pm0.02^{\dagger}$ & $0.99\pm0.02^{\dagger}$ & $0.99\pm0.02^{\dagger}$ \\
\hline
\multirow{6}*{ColonBreast} & SAM & $0.98\pm0.08$ & $0.98\pm0.08$ & $0.97\pm0.06$ & $0.98\pm0.05^{\dagger}$ & $0.99\pm0.03^{\dagger}$ & $0.97\pm0.07$ & $0.97\pm0.06$ & $0.97\pm0.06$ & $0.97\pm0.06$ \\
\cline{2-11}
 & GeoDE & $0.99\pm0.04^{\dagger}$ & $0.99\pm0.04^{\dagger}$ & $0.99\pm0.04^{\dagger}$ & $0.98\pm0.05$ & $0.95\pm0.08$ & $0.95\pm0.08$ & $0.95\pm0.08$ & $0.95\pm0.08$ & $0.98\pm0.05$ \\
 \cline{2-11}
 & ReliefF & $0.95\pm0.08$ & $0.95\pm0.12$ & $0.95\pm0.11$ & $0.94\pm0.12$ & $0.98\pm0.05$ & $1.00\pm0.00^{\dagger}$ & $0.97\pm0.08$ & $0.96\pm0.08$ & $0.97\pm0.05$ \\
 \cline{2-11}
 & Info\_Gain & $0.98\pm0.05$ & $0.95\pm0.08$ & $0.95\pm0.08$ & $0.95\pm0.08$ & $0.98\pm0.08$ & $0.98\pm0.08$ & $0.99\pm0.04$ & $0.98\pm0.08$ & $0.95\pm0.12$ \\
 \cline{2-11}
 & Func\_Pert & $0.98\pm0.08$ & $0.99\pm0.04^{\dagger}$ & $0.98\pm0.05$ & $0.98\pm0.05$ & $0.97\pm0.06$ & $0.99\pm0.03$ & $0.99\pm0.03^{\dagger}$ & $0.98\pm0.05$ & $0.98\pm0.05$ \\
 \cline{2-11}
 & EFSIS & $0.98\pm0.08$ & $0.98\pm0.08$ & $0.99\pm0.04^{\dagger}$ & $0.98\pm0.05$ & $0.96\pm0.07$ & $0.98\pm0.05$ & $0.98\pm0.05$ & $0.99\pm0.04^{\dagger}$ & $0.99\pm0.04^{\dagger}$ \\
\hline
\end{tabular}}
\caption{Predictive performance of six rankers on six datasets with different percentages of selected features: mean AUC and standard deviation. The superscript dagger ($\dagger$) indicates the best ranker in one experiment (of one specific dataset and percentage of selected features), and the superscript star ($\ast$) and bold font indicate the rankers that are significantly worse than the best individual one.}
\end{table}

The mean AUC (averaging the AUCs from ten-fold cross-validation) and associated standard deviation of four individual rankers and two ensemble ones (basic function perturbation and EFSIS) tested on 6 datasets with 9 different percentages of selected features are shown in Table 2. For each combination of dataset and percentage of selected features, the best ranker (the one with the highest mean AUC and lowest standard deviation) is marked with dagger, and the ones that are significantly worse than the best one are marked with star and are in bold font ($P$-value $<0.05$, Wilcoxon Signed-Ranks Test \cite{Demsar2006}). It shows a problem of the individual rankers: some individual rankers perform quite well in some datasets but poorly in some others. For example, GeoDE performs quite well in dataset CNS (it achieves the highest prediction accuracy among all rankers 7 times out of 9), but performs unsatisfactorily in dataset DLBCL (it achieves a significantly lower prediction accuracy than the best one 8 times out of 9, which makes it the worst for this dataset). But ReliefF performs contrarily to GeoDE in these two datasets. Since the performance of feature selection methods varies from dataset to dataset, it is difficult for researchers to choose an adequate one for their dataset.

The results in Table 2 show that the predictive performance of ensemble rankers is more stable across the different datasets analyzed. Function perturbation and EFSIS are significantly worse than the best ranker in 4 out of the 54 experiments (6 datasets $\times$ 9 percentages of selected features), while individual rankers are worse in 8, 10, 6, 7 experiments, respectively.

\subsection{Experimental results of stability performance}

\begin{figure}[ht]
\centering\includegraphics[width=1\linewidth]{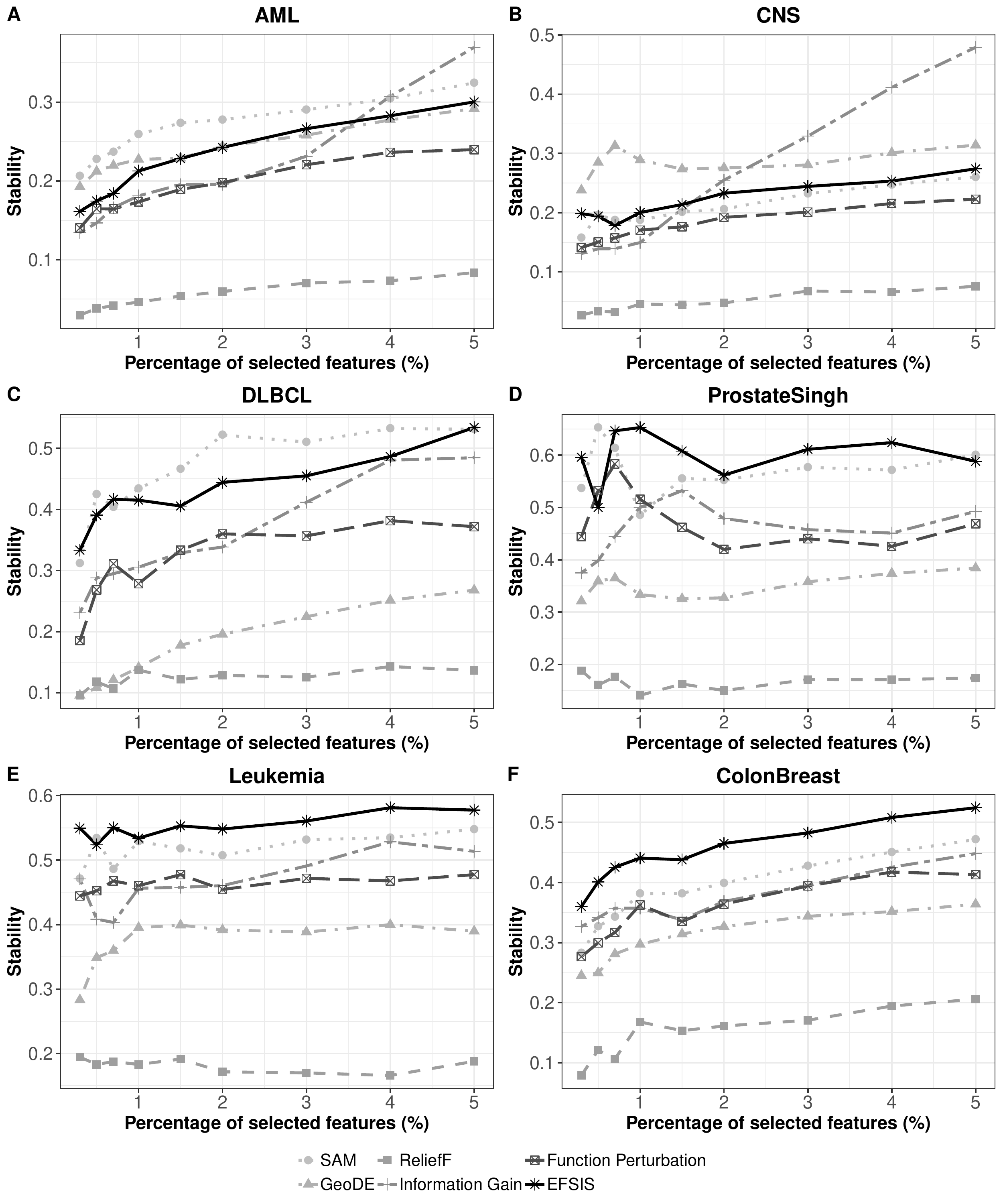}
\caption{Stability performance of six rankers on six datasets, tested with different percentages of selected features. For each dataset, four individual rankers (SAM, GeoDE, ReliefF, Information Gain), basic Function Perturbation, and EFSIS are considered.}
\end{figure}

\begin{figure}[ht]
\centering\includegraphics[width=0.5\linewidth]{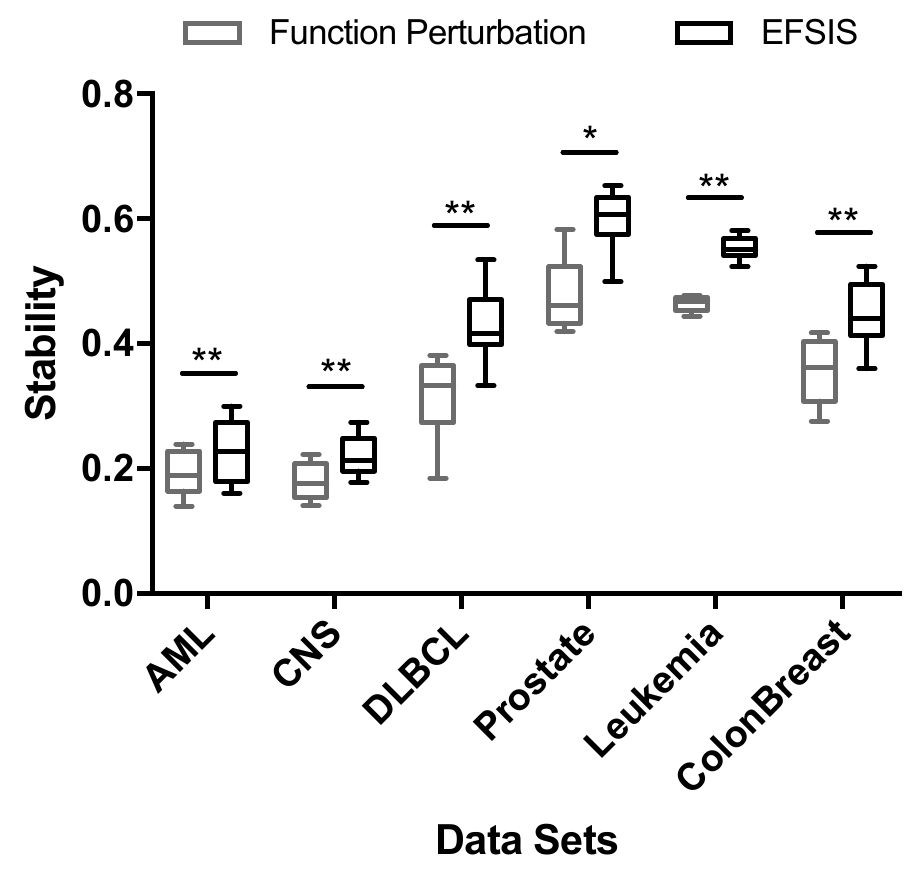}
\caption{Comparison of basic Function Perturbation and EFSIS in stability performance on six datasets. ** $=P$-value $<0.005$, * $=P$-value $<0.01$.}
\end{figure}

In this section, we will study the stability of the rankers. In the same way as in the evaluation of predictive performance, the stability was tested with different percentages of selected features on 6 datasets. 

Figure 2 shows the performance of the four individual rankers and the two ensemble rankers. Let us firstly look at the individual ones. GeoDE has the same problem as in the previous section: it achieves a very high stability in the CNS dataset but a very low one in the DLBCL dataset. ReliefF seems to be a very unstable method with the lowest stability score across all datasets, even in the dataset DLBCL where it showed great predictive performance (as mentioned in the previous section). 

When we compare basic function perturbation with the four individual rankers across the 6 datasets as shown in Figure 2, we can find that basic function perturbation performs moderately: it is never the best neither the worst compared with the individual rankers. EFSIS is even though not the best one in the first 3 datasets (Figure 2 A-C), but it performs better than all the individual rankers in the latter 3 datasets (Figure 2 D-F). If we compare between basic function perturbation and EFSIS which combines data perturbation and function perturbation, Figure 2 shows clearly that EFSIS consistently improves the stability of basic function perturbation. The box plot in Figure 3 shows the comparison between these two ensemble rankers on 6 datasets with the star ($\ast$) indicating the significance of difference ($P$-value was calculated using Wilcoxon Signed-Ranks Test \cite{Demsar2006}). We can see that the stability of EFSIS is significantly higher than basic function perturbation in all 6 datasets.

\section{Discussion and conclusions}
We have described a new framework for ensemble feature selection, which combines function perturbation and data perturbation and utilizes the stability of the individual methods as weight. The new method possesses the advantages of function perturbation and data perturbation: it combines the results from different individual feature selection methods and shows robust predictive performance, and it also provides more stable selected feature subsets. Therefore it frees the researchers from choosing the most suitable feature selection method for their datasets. Also, compared to basic function perturbation, it provides higher stability.

A major shortcoming of EFSIS, however, is that it is more time-consuming and more computationally expensive compared to the other methods assessed here. However it can be sped up by parallel computing. The parallelisation can be done in multiple ways. What we have tried was to split the jobs by individual rankers so that the job corresponding to one ranker was performed by one node. The time then depends on the slowest method since the aggregation needs the results of all individual methods. Parallelisation can considerably shorten the computing time, but depends on available computing resources.

Our work is the first study, to our knowledge, exploring the combination of function and data perturbation. It can form the basis for further studies in this direction. In the EFSIS framework, we have chosen to perform data perturbation in the first phase so that each ranker (feature selection method) is performed on all bootstrap datasets to produce one ranking that is next combined with rankings from the other rankers. In this way we can obtain the stability of each individual ranker based on the same subsets of samples, enabling us to use the stability estimates when combining results across the rankers. However, it would be interesting to explore an alternative approach where function perturbation is applied to each bootstrap dataset, which will produce $M$ ranked lists. In the next step, these $M$ lists will be combined (using rank product) to obtain the final ranked list. The idea behind this strategy is to make use of data perturbation's ability to improve the stability of function perturbation. Future studies will include these and other directions.

\end{document}